\documentclass{article}

\usepackage{times}
\usepackage{epsfig}
\usepackage{graphicx}
\usepackage{amsmath}
\usepackage{amssymb}
\usepackage{tabularx}
\usepackage{mathtools}
\usepackage{algorithm}
\usepackage{algpseudocode}
\usepackage{amsmath}
\usepackage{bm}

\usepackage[dvipsnames]{xcolor}

\usepackage{graphicx}
\usepackage{booktabs}
\usepackage[lofdepth,lotdepth]{subfig}
\usepackage{appendix}
\usepackage{multirow}

\PassOptionsToPackage{numbers, compress}{natbib}
\usepackage[final]{neurips_2022}

\usepackage[utf8]{inputenc} 
\usepackage[T1]{fontenc}    
\usepackage{hyperref}       
\usepackage{url}            
\usepackage{booktabs}       
\usepackage{amsfonts}       
\usepackage{nicefrac}       
\usepackage{microtype}      
\usepackage{xcolor}         

\title{How Good Are Deep Generative Models for Solving Inverse Problems?
}

\author{
Shichong Peng, Alireza Moazeni, Ke Li \\
APEX Lab \\
School of Computing Science \\
Simon Fraser University\\
{\tt\small \{shichong\_peng,seyed\_alireza\_moazenipourasil,keli\}@sfu.ca}}

\begin{document}

\maketitle

\begin{abstract}

  Deep generative models, such as diffusion models, GANs, and IMLE, have shown impressive capability in tackling inverse problems. However, the validity of model-generated solutions w.r.t. the forward problem and the reliability of associated uncertainty estimates remain understudied. This study evaluates recent diffusion-based, GAN-based, and IMLE-based methods on three inverse problems, i.e., $16\times$ super-resolution, colourization, and image decompression. 
  We assess the validity of these models' outputs as solutions to the inverse problems and conduct a thorough analysis of the reliability of the models' estimates of uncertainty over the solution. 
  Overall, we find that the IMLE-based CHIMLE method outperforms other methods in terms of producing valid solutions and reliable uncertainty estimates.
\end{abstract}

\section{Introduction}

Deep generative models have seen remarkable growth in adoption across various applications. These models, exemplified by methods such as diffusion models~\cite{sohl2015deep,ho2020denoising}, Generative Adversarial Networks (GANs)~\cite{goodfellow2014generative}, and Implicit Maximum Likelihood Estimation (IMLE)~\cite{li2018implicit}, have demonstrated remarkable capabilities in capturing complex data distributions. 

One particularly intriguing application of deep generative models is solving inverse problems, a class of challenges that involves reconstructing clean data with complex structure, often from incomplete or noisy observations. Instead of producing a single deterministic solution, these models can learn entire distributions over the possible solutions. This capability is pivotal when the forward problem is not injective, in which case there is no unique solution, but rather many solutions. In such cases, having both a clean solution and a quantification of the uncertainty in the solution are important. 

However, a critical aspect often overlooked in the evaluation of deep generative models is the validity of their outputs as solutions to the underlying inverse problem. More concretely, a valid solution is one where if the forward problem were applied to the solution, the result would be similar with the provided input. In other words, while deep generative models excel at generating outputs that appear plausible, they may not actually be valid solutions to the inverse problem. This validation is of great importance, since otherwise the generated solutions would not truly invert the forward problem.

Since the forward problem is often not injective, another dimension of evaluation that deserves more attention is the accuracy of uncertainty estimates. Accurate pixel-level modelling of uncertainty in the generated solution is crucial, particularly in domains where decisions carry significant consequences. 

In this study, we conduct an evaluation of recent deep generative models based on a variety of methods, including diffusion, GAN, and IMLE, in the context of three challenging inverse problems: $16\times$ single image super-resolution, image colourization and image decompression. Our evaluation focuses on two critical dimensions that are traditionally underemphasized: output validity and model uncertainty. We aim to assess the extent to which these models' solutions align with the input data when passed through the forward problem, and to quantify and analyze the inherent uncertainty in their solutions.

\section{Background}

\subsection{Deep Generative Models for Inverse Problems}

Given the input $\mathbf{x}$ and a forward problem $\mathcal{F}$, the goal of inverse problems is to find solutions $\mathbf{y}$ such that $\mathcal{F}(\mathbf{y}) = \mathbf{x}$.
When $\mathcal{F}$ is not injective, there can be multiple plausible solutions $\mathbf{y}$ for the same observed data $\mathbf{x}$, i.e., $\mathbf{y} \in \{\mathbf{y}' \vert \mathcal{F}(\mathbf{y}') = \mathbf{x}\}$. In light of this, deep generative models estimate $p(\mathbf{y} | \mathbf{x})$ to capture the distribution of all plausible solutions.

When $\mathcal{F}$ is additive Gaussian, one way to model it is with diffusion models~\cite{sohl2015deep,ho2020denoising}. Specifically, diffusion models break down $\mathcal{F}$ into $T$ additive Gaussian steps that progressively add noise to the data according to a predefined variance schedule. If we denote the original data as $\mathbf{x}_0$, then each step of the forward process $q(\mathbf{x}_{1:T}|\mathbf{x}_{0})$ can be written as:
\begin{align}
    q(\mathbf{x}_{1:T}|\mathbf{x}_{0}) \coloneqq \prod_{t=1}^{T} q(\mathbf{x}_{t}|\mathbf{x}_{t-1}),\ \ \  q(\mathbf{x}_{t}|\mathbf{x}_{t-1}) \coloneqq \mathcal{N}(\mathbf{x}_t; \sqrt{1-\beta_{t}}\mathbf{x}_{t-1},\beta_t\mathbf{I})
\end{align}
Here, $\beta_1,\dots,\beta_T$ represent the variance schedule. To model the reverse process $p_{\theta}(\mathbf{x}_{0:T})$, diffusion models make an additional assumption, namely that the reverse process is a Markov chain with Gaussian transition kernels, starting at $p(\mathbf{x}_T)=\mathcal{N}(\mathbf{x}_T; 0,\mathbf{I})$:   
\begin{align}
    p_{\theta}(\mathbf{x}_{0:T}) \coloneqq p(\mathbf{x}_T) \prod_{t=1}^{T} p_{\theta}(\mathbf{x}_{t-1}|\mathbf{x}_{t}),\ \ \  p_{\theta}(\mathbf{x}_{t-1}|\mathbf{x}_{t}) \coloneqq \mathcal{N}(\mathbf{x}_{t-1}; \mathbf{\mu}_{\theta}(\mathbf{x}_t, t),\mathbf{\Sigma}_{\theta}(\mathbf{x}_t, t))
\end{align}
where $\mathbf{\mu}_{\theta}(\mathbf{x}_t, t),\mathbf{\Sigma}_{\theta}(\mathbf{x}_t, t)$ are the predicted mean and covariance at time step $t$. It's worth noting that the reverse process is only truly the inverse of the forward process when there are an infinite number of time steps $T$; otherwise the Gaussianity assumption in the transition kernel of the reverse process would not hold. In practice, a finite $T$ is used, in which case the Gaussianity assumption would introduce approximation errors.
Therefore, the assumptions of diffusion models are technically not met when the forward problem is not additive Gaussian or when the number of time steps is small.

One alternative is to use a conditional GAN, comprising a generator and a discriminator. The generator $G_\theta$ takes in the observed input $\mathbf{x}$ and a latent code $\mathbf{z} \sim \mathcal{N}(0,\mathbf{I})$ and generates an output $\mathbf{y} \coloneqq G_{\theta}(\mathbf{x},\mathbf{z})$. The discriminator's role is to distinguish between the generated sample and real data example. However, due to mode collapse, the generator tends to generate identical samples for the same input and ignores the latent noise~\cite{isola2017image}. 
To mitigate this problem, various approaches have been proposed, including introducing additional losses in the latent space~\cite{zhu2017toward}, adding mode-seeking terms~\cite{Mao2019ModeSG}, or using contrastive losses~\cite{Liu2021DivCoDC,zhan2022monce,hu2022qs}. While these methods succeed in enhancing sample diversity, they often come at the expense of reduced sample fidelity.

Another alternative approach is Implicit Maximum Likelihood Estimation (IMLE)~\citep{li2018implicit}. IMLE differs from GANs in two key ways: it explicitly aims to cover all modes, and optimizes a non-adversarial objective. To achieve the former, IMLE reverses the direction in which generated samples are matched to real data: rather than making each generated sample similar to some real data point, it makes sure each real data point has a similar generated sample. To achieve the latter, it removes the discriminator (which matches generated samples to real data implicitly) and instead explicitly performs matching using nearest neighbour search. The latter can be done efficiently using DCI~\citep{li2016fast,li2017fast}, which avoids the curse of dimensionality. Recent advancements to IMLE include cIMLE~\cite{Li2020MultimodalIS}, which extends IMLE to the conditional setting, and CHIMLE~\cite{peng2022chimle}, which further enhances generated image quality by adopting a hierarchical approach to constructing latent codes.

\subsection{Model Uncertainty Quantification}

A popular and statistically rigorous approach to perform uncertainty quantification is conformal prediction~\cite{Bates2021DistributionFreeRP,Angelopoulos2021AGI,Einbinder2022ConformalPI,Teng2022PredictiveIW}. Conformal prediction stands out for its ability to construct statistically valid uncertainty guarantees for general predictors without relying on distributional or model assumptions. Recent studies~\cite{Angelopoulos2022ImagetoImageRW,Teneggi2023HowTT,Horwitz2022ConffusionCI} have extended the application of conformal prediction to address image-to-image translation problems. In this study, we adopt a sampling-based approach, as outlined in \cite{Horwitz2022ConffusionCI}, to evaluate model uncertainty across different methods.

\section{Experiments}

We evaluate recent deep generative models for inverse problems that are based on diffusion, GAN and IMLE. For diffusion-based methods, we choose DDRM~\cite{kawar2022denoising} and NDM~\cite{Batzolis2021ConditionalIG}. For GAN-based methods, we choose BicyleGAN~\cite{Zhu2017TowardMI}, MSGAN~\cite{Mao2019ModeSG}, DivCo~\cite{Liu2021DivCoDC} and MoNCE~\cite{zhan2022monce}. For IMLE-based methods, we choose cIMLE~\cite{Li2020MultimodalIS} and CHIMLE~\cite{peng2022chimle}. 

Our evaluation includes three challenging inverse problems: $16\times$ single image super-resolution, image colourization and image decompression. For image super-resolution, we choose three categories from ILSVRC-2012~\cite{Russakovsky2015ImageNetLS} with an input resolution of $32\times32$, and output resolution of $512\times512$. For image colourization, we choose two categories from ILSVRC-2012~\cite{Russakovsky2015ImageNetLS} and the Natural Color Dataset (NCD)~\cite{Anwar2020ImageCA}. For image decompression, we choose the RAISE1K~\cite{dang2015raise} dataset and compressed each image using JPEG with a quality of 1\%.

\begin{table}[t]
    \vspace{-2em}
    \centering
    \footnotesize
    \resizebox{\linewidth}{!}{
    \begin{tabular}{lccccccccc}
    \toprule
     & \multicolumn{3}{c}{Super-Resolution (SR)} & \multicolumn{3}{c}{Colourization (Col)} & \multicolumn{3}{c}{Image Decompression (DC)} \\
    \midrule
     & \textit{LPIPS} $\downarrow$  & \textit{PSNR} $\uparrow$ & \textit{SSIM} $\uparrow$ & \textit{LPIPS} $\downarrow$  & \textit{PSNR} $\uparrow$ & \textit{SSIM} $\uparrow$ & \textit{LPIPS} $\downarrow$  & \textit{PSNR} $\uparrow$ & \textit{SSIM} $\uparrow$\\
    \midrule
    \multicolumn{7}{l}{\textit{GAN-based:}}\\
    \ \textit{BicyleGAN~\cite{Zhu2017TowardMI}}  & $0.105$ & $22.31$ & $0.832$ & $0.322$ & $20.28$ & $0.716$ & $0.349$ & $19.75$ & $0.781$  \\
    \ \textit{MSGAN~\cite{Mao2019ModeSG}}  & $0.135$ & $20.83$ & $0.810$ & $0.370$ & $18.98$ & $0.665$ & $0.416$ & $17.24$ & $0.712$ \\
    \ \textit{DivCo~\cite{Liu2021DivCoDC}} & $0.136$ & $20.60$ & $0.763$ & $0.336$ & $18.84$ & $0.687$  & $0.356$ & $18.94$ & $0.731$ \\
    \ \textit{MoNCE~\cite{zhan2022monce}} & $0.091$ & $27.41$ & $0.943$ & $0.030$ & \underline{$39.52$} & \boldsymbol{$0.990$} & \underline{$0.213$} & \underline{$25.78$} & \underline{$0.854$} \\
    \multicolumn{7}{l}{\textit{Diffusion-based:}}\\
    \ \textit{DDRM~\cite{kawar2022denoising}}  & $0.045$ & \underline{$29.57$} & $0.948$ & $0.127$ & $28.21$ & $0.922$ & $0.539$ & $19.07$ & $0.459$ \\
    \ \textit{NDM~\cite{Batzolis2021ConditionalIG}}  & $0.082$ & $23.50$ & $0.854$ & $0.057$ & $37.34$ & $0.951$ & $0.525$ & $14.15$ & $0.575$ \\
    \multicolumn{7}{l}{\textit{IMLE-based:}}\\
    \ \textit{cIMLE~\cite{Li2020MultimodalIS}} & \underline{$0.040$} & $28.57$ & \underline{$0.949$} & \underline{$0.022$} & $36.12$ & \underline{$0.981$} & $0.261$ & $22.39$ & $0.790$  \\
    \ \textit{CHIMLE~\cite{peng2022chimle}}  & \boldsymbol{$0.009$} & \boldsymbol{$33.26$}  & \boldsymbol{$0.988$} & \boldsymbol{$0.011$} & \boldsymbol{$41.25$} & \boldsymbol{$0.990$} & \boldsymbol{$0.191$} & \boldsymbol{$26.24$} & \boldsymbol{$0.877$} \\
    
    \bottomrule
  \end{tabular}
  }  \caption{Comparison of model output validity among diffusion-based, GAN-based and IMLE-based methods. The validity is measured by computing the LPIPS~\cite{zhang2018unreasonable}, PSNR and SSIM between the model input image and the solution to the forward problem applied to the generated samples, averaged over 50 runs. Lower values of LPIPS are better and higher values of PSNR and SSIM are better. 
  Notably, CHIMLE~\cite{peng2022chimle} consistently achieves the best output validity in our evaluation.
  }
    \label{tab:consistency}
\end{table}

\subsection{Output Validity Assessment}
The first evaluation focuses on assessing the validity of the model output as a solution to the inverse problem. Recall that in an inverse problem, it is essential that the provided input is similar to the solution to the forward problem with the model output as input. 
To assess this, we generate 50 random samples for each input image in the test set. These samples are then used as inputs to solve the corresponding forward problem for each task, namely bicubic downsampling by a factor of $16\times$ for super-resolution, conversion of RGB images to grayscale for colourization, and compression of images using JPEG at a quality of 1\% for image decompression. Subsequently, we compare the processed outputs to the original input.

Table~\ref{tab:consistency} shows the average LPIPS~\cite{zhang2018unreasonable}, PSNR and SSIM metrics between the model input and the solution to the forward problem applied to the model output. Lower LPIPS distances are better, whereas higher PSNR and SSIM are better. 
We found that  CHIMLE~\cite{peng2022chimle}, an IMLE-based method, performs the best across tasks and metrics. In second place, we have cIMLE~\cite{Li2020MultimodalIS} according to LPIPS and SSIM on super-resolution and colourization, DDRM~\cite{kawar2022denoising} according to PSNR on super-resolution, and MoNCE~\cite{zhan2022monce} according to PSNR on colourization and according to all metrics on decompression. 

\begin{figure}[ht]
    \centering
    \vspace{-2em}
  \subfloat[Input]{
    \includegraphics[width=0.19\linewidth]{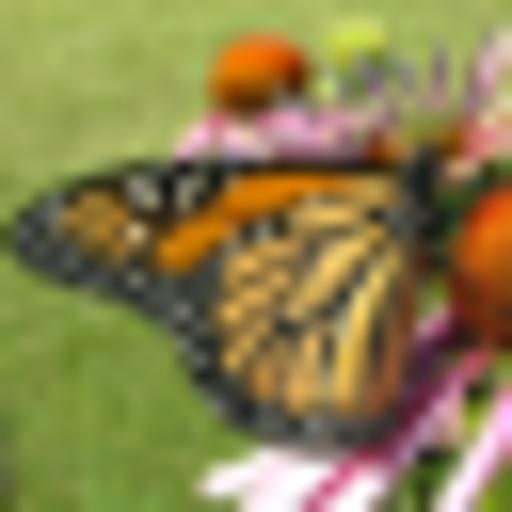}
  }
  \subfloat[BicycleGAN~\cite{Zhu2017TowardMI}]{
    \includegraphics[width=0.19\linewidth]{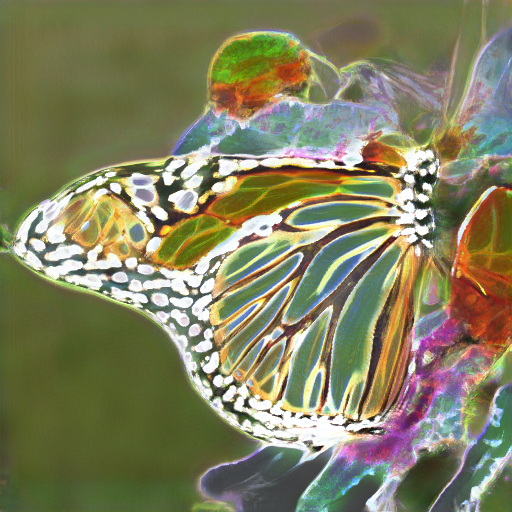}
  }
  \subfloat[MSGAN~\cite{Mao2019ModeSG}]{
    \includegraphics[width=0.19\linewidth]{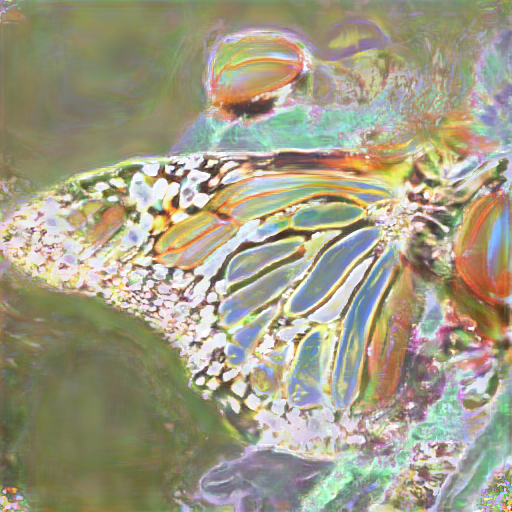}
  }
  \subfloat[DivCo~\cite{Liu2021DivCoDC}]{
    \includegraphics[width=0.19\linewidth]{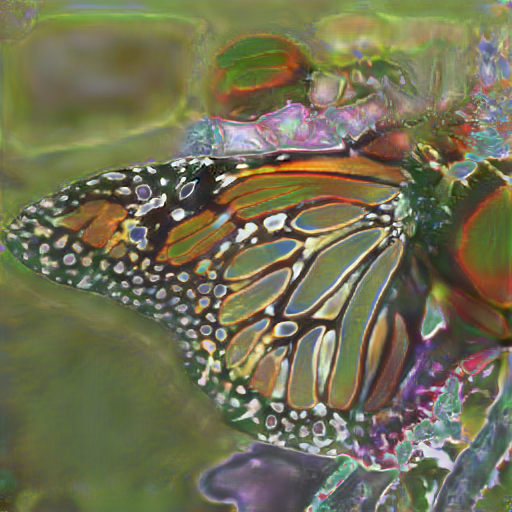}
  }
  \subfloat[MoNCE~\cite{zhan2022monce}]{
    \includegraphics[width=0.19\linewidth]{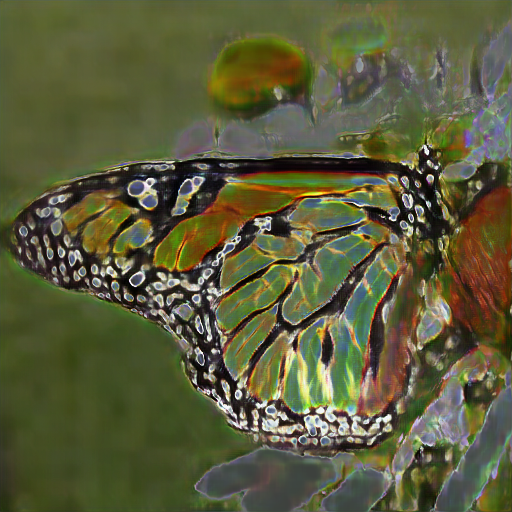}
  }\\
  \subfloat[DDRM~\cite{kawar2022denoising}]{
    \includegraphics[width=0.19\linewidth]{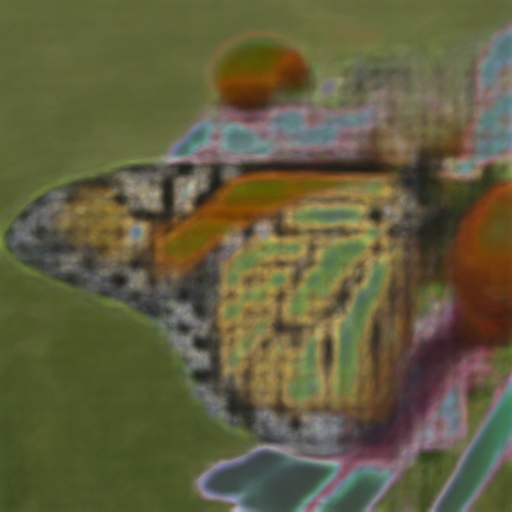}
  }
  \subfloat[NDM~\cite{Batzolis2021ConditionalIG}]{
    \includegraphics[width=0.19\linewidth]{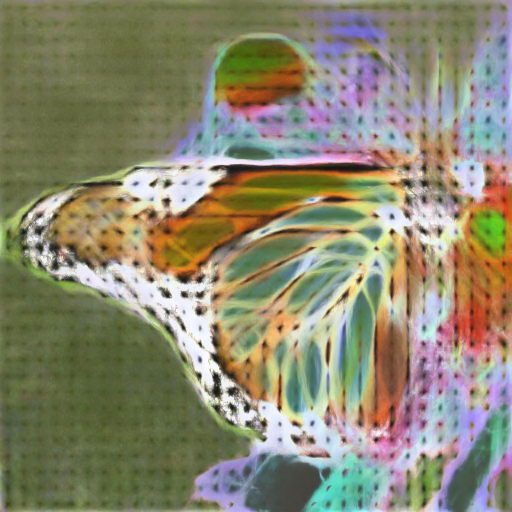}
  }
  \subfloat[cIMLE~\cite{Li2020MultimodalIS}]{
    \includegraphics[width=0.19\linewidth]{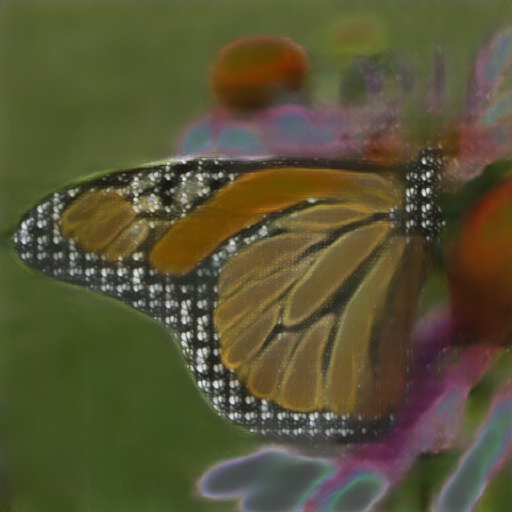}
  }
  \subfloat[CHIMLE~\cite{peng2022chimle}]{
    \includegraphics[width=0.19\linewidth]{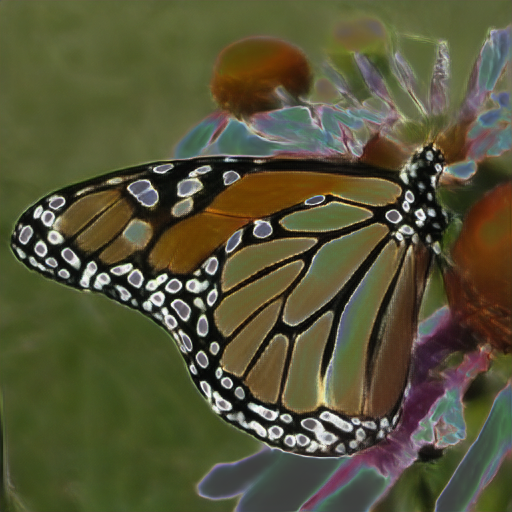}
  }
  \subfloat[Original Image]{
    \includegraphics[width=0.19\linewidth]{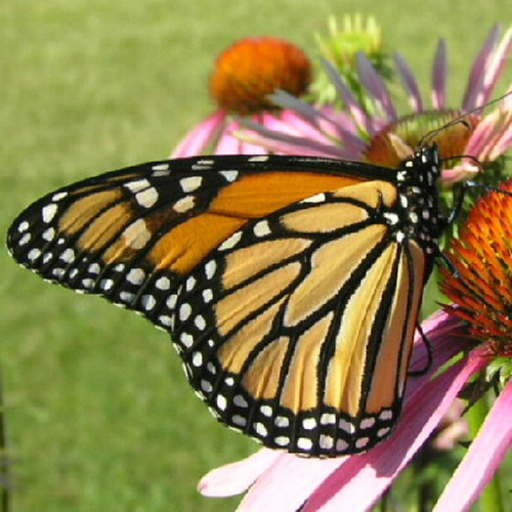}
  }
  \caption{Comparison of model uncertainty visualized as confidence intervals extracted from 50 randomly generated samples from each method using the same input. As shown, CHIMLE~\cite{peng2022chimle} produces high-fidelity outputs and uncertainty estimates that closely align with ambiguous regions in the input (e.g., around the black rim of the butterfly's wing). 
  In contrast, other methods either produce low fidelity results (DivCo~\cite{Liu2021DivCoDC}, MoNCE~\cite{zhan2022monce}, DDRM~\cite{kawar2022denoising} and cIMLE~\cite{Li2020MultimodalIS}) or contain excessive uncertainty in less ambiguous input regions such as the thick black stripes in the wing's center (BicycleGAN~\cite{Zhu2017TowardMI}, MSGAN~\cite{Mao2019ModeSG} and NDM~\cite{Batzolis2021ConditionalIG}).   }
  \label{fig:uncertainty}
\end{figure}

\subsection{Model Uncertainty Evaluation}
To evaluate model uncertainty for each method, we adopt a sampling-based approach mentioned in \cite{Horwitz2022ConffusionCI}. For each input, we randomly generate 50 samples from each model and calculate the confidence interval at each pixel. In our visualization, the confidence interval is superimposed as highlights on top of the dimmed mean generated sample. A wider confidence interval corresponds to greater model uncertainty, visually appeared as a brighter colour. To determine the size of the confidence interval, we subtract the lower quantile from the upper quantile for each pixel within the set of generated samples for the same input. In our evaluation, we set the lower quantile at the $5^{th}$ percentile and the upper quantile at the $95^{th}$ percentile.
Additionally, when visualizing the confidence interval for a specific input, we calibrate the lower and upper bounds using the remaining images in the test set, with more details available in \cite{Horwitz2022ConffusionCI}.

Figure~\ref{fig:uncertainty} shows the model uncertainty comparison in the task of $16\times$ image super-resolution. 
As shown, BicycleGAN~\cite{Zhu2017TowardMI}, MSGAN~\cite{Mao2019ModeSG} and NDM~\cite{Batzolis2021ConditionalIG} generate samples with significant variations. However, these methods also show substantial uncertainty in regions that are constrained significantly by the input and therefore should not be uncertain, e.g., the thick black strips in the wing's center should be just solid black with no other possibilities. On the other hand, DDRM~\cite{kawar2022denoising} generates blurry outputs and shows low uncertainty in regions where the input is ambiguous, where one would expect a higher degree of uncertainty, e.g., along the rim of the butterfly's wing. DivCo~\cite{Liu2021DivCoDC}, MoNCE~\cite{zhan2022monce}, and cIMLE~\cite{Li2020MultimodalIS} show model uncertainties that align with the degree of ambiguity present in the input, but their outputs are low in visual fidelity. CHIMLE~\cite{peng2022chimle} achieves the overall best performance and produces realistic outputs and uncertainty estimates that accurately capture input ambiguity. Visual comparisons of model uncertainty for the other tasks, namely image colourization and image decompression, can be found in the appendix.

\section{Conclusion}
In this study, we thoroughly evaluated diffusion-based, GAN-based, IMLE-based methods on three challenging inverse problems, namely $16\times$ super-resolution, colourization, and image decompression. Our assessment focused on two critical dimensions: output validity and model uncertainty. Surprisingly, despite their popularity, diffusion-based methods do not perform well on either dimension, whereas GAN-based and IMLE-based methods perform better. Among GAN-based and IMLE-based methods, CHIMLE~\cite{peng2022chimle} seems to perform the best along both dimensions.

\clearpage
\newpage

\bibliography{neurips_2022}
\bibliographystyle{plain}

\clearpage
\newpage
\appendix

\section{Additional Results on Model Uncertainty}

\begin{figure}[ht]
    \centering

  \subfloat[Input]{
    \includegraphics[width=0.19\linewidth]{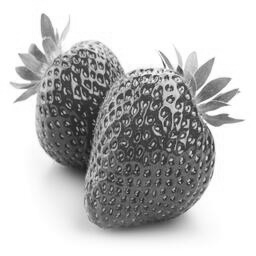}
  }
  \subfloat[BicycleGAN]{
    \includegraphics[width=0.19\linewidth]{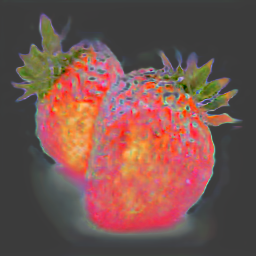}
  }
  \subfloat[MSGAN]{
    \includegraphics[width=0.19\linewidth]{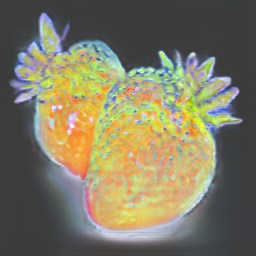}
  }
  \subfloat[DivCo]{
    \includegraphics[width=0.19\linewidth]{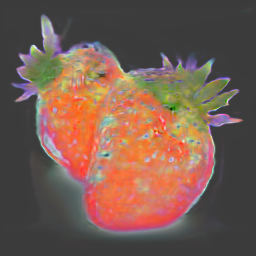}
  }
  \subfloat[MoNCE]{
    \includegraphics[width=0.19\linewidth]{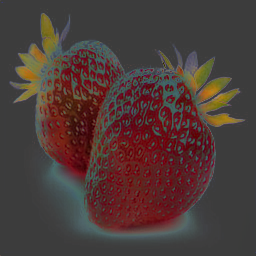}
  }\\
  \subfloat[DDRM]{
    \includegraphics[width=0.19\linewidth]{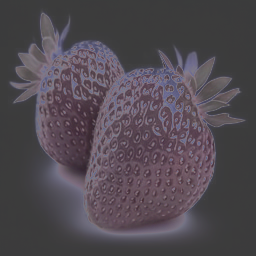}
  }
  \subfloat[NDM]{
    \includegraphics[width=0.19\linewidth]{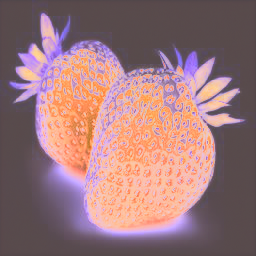}
  }
  \subfloat[cIMLE]{
    \includegraphics[width=0.19\linewidth]{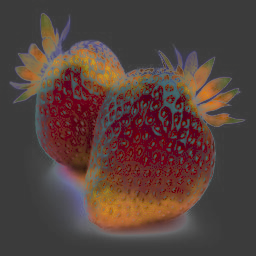}
  }
  \subfloat[CHIMLE]{
    \includegraphics[width=0.19\linewidth]{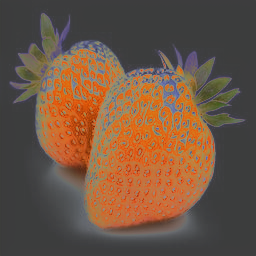}
  }
  \subfloat[Original Image]{
    \includegraphics[width=0.19\linewidth]{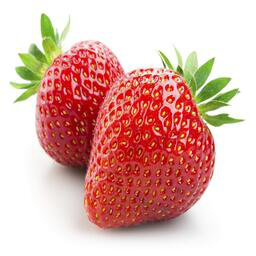}
  }
  \caption{Model uncertainty comparison on image colourization.}
  \label{fig:col}
\end{figure}

Figure~\ref{fig:col} shows the model uncertainty comparison of different methods on the task of image colourization. In our visualization, the confidence interval is superimposed as highlights on top of the dimmed mean generated sample. A wider confidence interval corresponds to greater model uncertainty, visually appeared as a brighter colour.
As shown, BicycleGAN, MSGAN and DivCo produce diverse results but their samples fail to preserve the details from the input and lack visual fidelity (i.e., look unrealistic). 
Conversely, MoNCE generates samples with limited diversity, with uncertainty primarily concentrated in less ambiguous areas, such as the strawberry's leaf (where the colour of the strawberry leaf should be green). DDRM produces samples that show desaturation and minimal diversity, and fails to capture different plausible solutions (e.g., the strawberry's colour could be either red or green). NDM provides diverse samples but shows high uncertainty in regions that should not contain much diversity (e.g., the leaf on the strawberry). cIMLE's model uncertainty does not fully encompass the entire strawberry. In contrast, CHIMLE generates diverse samples while assigning low model uncertainty to regions in the input that are less ambiguous and high uncertainty to regions that should contain more diversity, capturing a comprehensive and precise representation of uncertainty in the input data.

\begin{figure}[ht]
    \centering

  \subfloat[Input]{
    \includegraphics[width=0.19\linewidth]{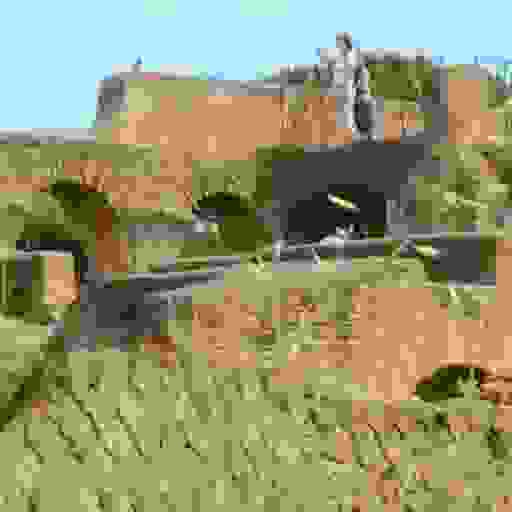}
  }
  \subfloat[BicycleGAN]{
    \includegraphics[width=0.19\linewidth]{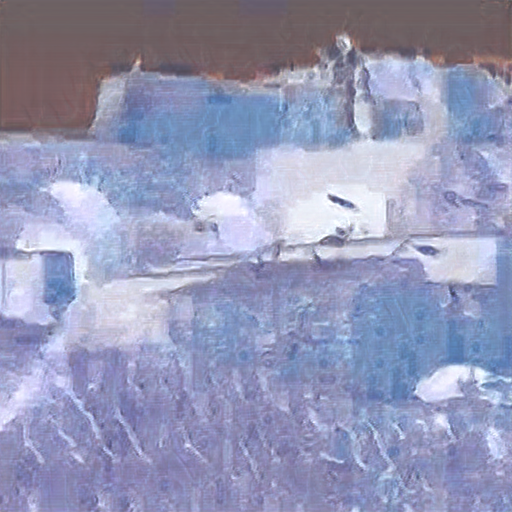}
  }
  \subfloat[MSGAN]{
    \includegraphics[width=0.19\linewidth]{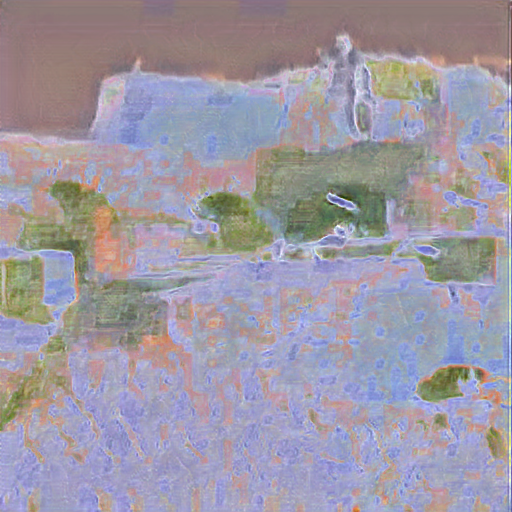}
  }
  \subfloat[DivCo]{
    \includegraphics[width=0.19\linewidth]{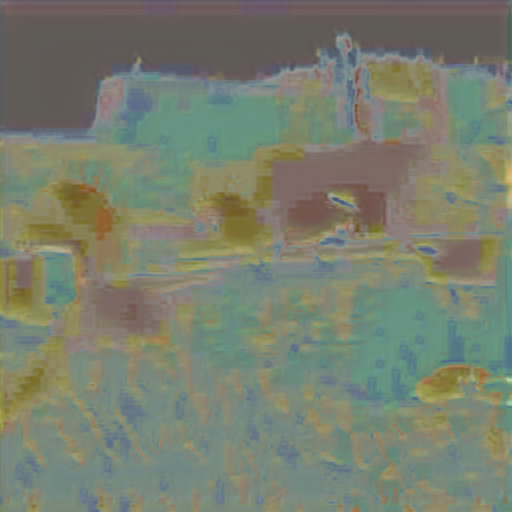}
  }
  \subfloat[MoNCE]{
    \includegraphics[width=0.19\linewidth]{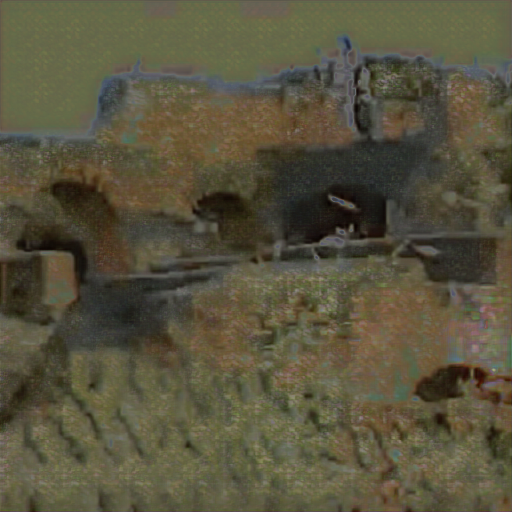}
  }\\
  \subfloat[DDRM]{
    \includegraphics[width=0.19\linewidth]{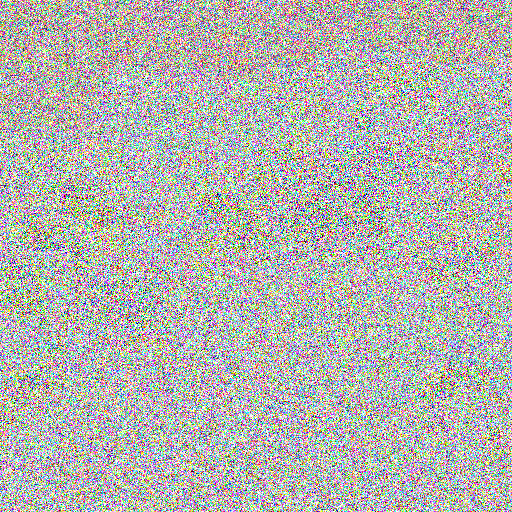}
  }
  \subfloat[NDM]{
    \includegraphics[width=0.19\linewidth]{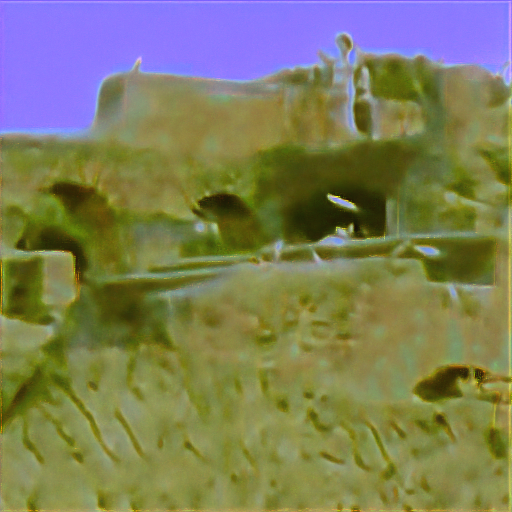}
  }
  \subfloat[cIMLE]{
    \includegraphics[width=0.19\linewidth]{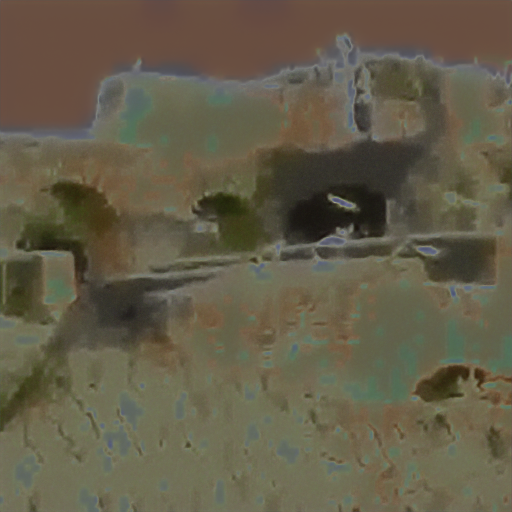}
  }
  \subfloat[CHIMLE]{
    \includegraphics[width=0.19\linewidth]{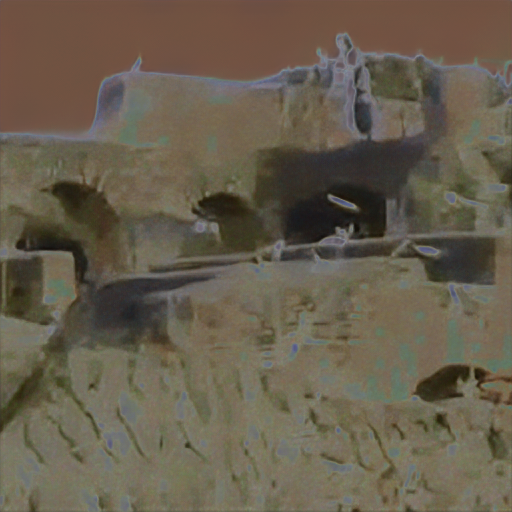}
  }
  \subfloat[Original Image]{
    \includegraphics[width=0.19\linewidth]{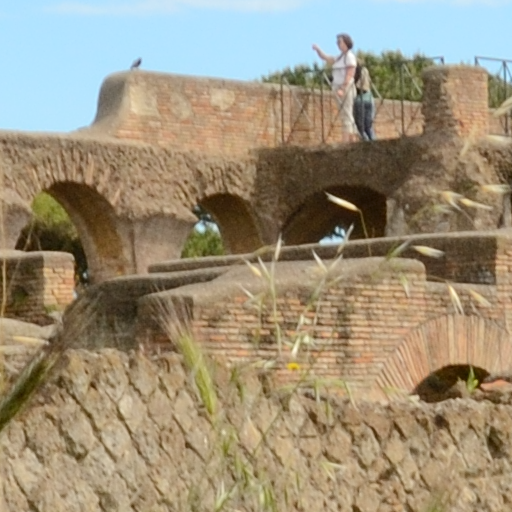}
  }
  \caption{Model uncertainty comparison on image decompression.}
  \label{fig:dc}
\end{figure}

Figure~\ref{fig:dc} shows the model uncertainty comparison of different methods on the task of image decompression. As shown, BicycleGAN, MSGAN and DivCo produce excessively diverse samples, resulting in model uncertainty that spans across the entire image. Conversely, MoNCE lacks sample diversity and fails to capture the inherent ambiguity present in the input. DDRM struggles to generate reasonable samples in this example. NDM produces diverse samples but the generated samples are blurry. The IMLE-based methods, cIMLE and CHIMLE, show model uncertainties that effectively capture the ambiguous regions in the input, such as the edges of the human and the trees in the top right corner.

\end{document}